\documentclass{article} 
\usepackage{iclr2019_conference,times}

\usepackage{amsmath,amsfonts,bm}

\def\eqref#1{equation~\ref{#1}}

\def\1{\bm{1}}

\DeclareMathAlphabet{\mathsfit}{\encodingdefault}{\sfdefault}{m}{sl}
\SetMathAlphabet{\mathsfit}{bold}{\encodingdefault}{\sfdefault}{bx}{n}

\usepackage{hyperref}
\usepackage{url}
\usepackage[title]{appendix}
\usepackage{graphicx}
\usepackage{amsmath}
\usepackage{float}
\usepackage{subcaption}
\usepackage{siunitx}
\title{Learning Inward Scaled Hypersphere Embedding: Exploring Projections in Higher Dimensions}
\iclrfinalcopy

\author{Muhammad Kamran Janjua \\
School of Electrical Engineering and Computer Science\\
National University of Sciences and Technology\\
Islamabad, Pakistan \\
\texttt{mjanjua.bscs16seecs@seecs.edu.pk} \\
\AND
Shah Nawaz, Alessandro Calefati, Ignazio Gallo \\
Department of Theoretical and Applied Science \\
University of Insubria \\
Varese, Italy \\
\texttt{\{shah.nawaz,a.calefati,ignazio.gallo\}@uninsubria.it} 
}

\begin{document}

\maketitle

\begin{abstract}
Majority of the current dimensionality reduction or retrieval techniques rely on embedding the learned feature representations onto a computable metric space. Once the learned features are mapped, a distance metric aids the bridging of gaps between similar instances. Since the scaled projection is not exploited in these methods, discriminative embedding onto a hyperspace becomes a challenge. In this paper, we propose to inwardly scale feature representations in proportional to projecting them onto a hypersphere manifold for discriminative analysis. We further propose a novel, yet simpler, convolutional neural network based architecture and extensively evaluate the proposed methodology in the context of classification and retrieval tasks obtaining results comparable to state-of-the-art techniques.\footnote{The accompanying code will be released.} 
\end{abstract}

\section{Introduction}
\label{sec:intro}
In last few years, mainly due to the advances in convolutional neural networks the performance on tasks such as image classification~\citep{szegedy2015going}, cross-modal and uni-modal retrieval~\cite{wang2016learning,park2016image}, and face recognition and verification~\citep{calefati2018git,wen2016discriminative,deng2017marginal} has increased drastically. It has been observed that deeper architectures tend to provide better capabilities in terms of approximating any learnable function. A common observation is that deeper architectures (large number of parameters) can "learn" features at various levels of abstraction. However, it is a well explored problem that deeper architectures are more prone to overfitting than their shallower counterparts, thus hampering their generalization ability, furthermore they are computationally expensive. Majority of convolutional neural networks (CNNs) based pipelines follow the same structure i.e. alternating convolution and max pool layers, fully connected along with activation functions and dropout for regularization~\citep{jarrett2009best,szegedy2017inception,simonyan2014very}. Recently, the work in~\citep{springenberg2014striving} proposed an all convolutional neural network, an architecture based on just CNN layers.

Another major reason for this drastic growth is discriminative learning techniques~\citep{sun2014deep,schroff2015facenet,wen2016discriminative} aiming at embedding the learned feature representations onto a hyperspace, linear or quadratic in most cases. There are studies in literature~\citep{aggarwal2001surprising,beyer1999nearest} arguing that in higher dimensions when the data is projected onto an input space there is not much divergence in terms of distance ratio of the nearest and farthest neighbors to a given target and tends to be $\approx 1$. Due to this relative contrast of the distance to an input point cannot be discriminated effectively. It is important to note that since retrieval and search tasks tend to operate on higher dimensions, this phenomenon is valid for these problems as well. The works done by~\citep{nawaz2018revisiting,park2016image} for cross modal retrieval assert that Recall@K (a metric depending on Euclidean distance for similarity computation between feature representations) is not a competitive metric to evaluate the retrieval systems. Euclidean distance can be formulated as $L_2 = \sqrt{(x_i-x_j)^2+(y_i-y_j)^2}$ where $x_i, x_j$ and $y_i, y_j$ are two points in the input space. Surprisingly enough,~\citep{aggarwal2001surprising} argues that in $L_k$-norms, the meaningfulness in high dimensionality is not independent of value of $k$ with lower values of $k$ norms performing better than their greater value counterparts i.e. $L_2 < L_1$. The general formula of $L_k$ norm can be setup as $L_{k}(x,y) =\sum_{i=1}^d(\|x_i-y_i\|^k)^{1/k}$ for $k = 1,2,3,...,n$. The relation considers norms with $k = \frac{1}{2}, \frac{1}{3}...\frac{1}{n};  \forall  k <  1, n \in Z$, referred to as fractional norms. Although fractional norms do not necessarily follow the triangle inequality $L_{k}(x,z) \leq L_{k}(x,y) + L_{k}(y,z); \forall x,y,z \in X$ where X is the input space, they tend to provide better contrast than their integral counterparts in terms of relative distances between query points and target. 

\begin{figure}[t!]
\centering
\includegraphics[scale=0.85]{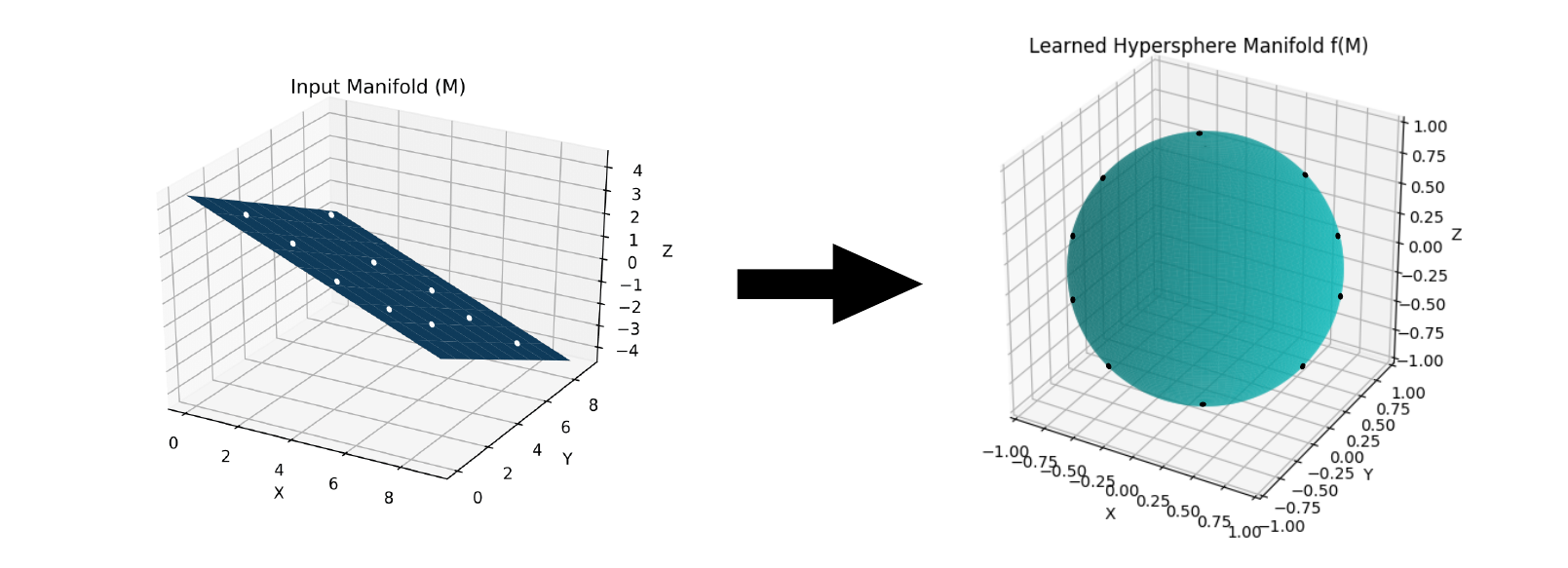}
\caption{A toy figure representing how the projection takes place. The manifold $M$ is transformed into a hypersphere $f(M)$ during training. The small black and white dots represent different classes. The figure is made with MNIST dataset in perspective. The perfect alignment of all the classes on the circumference of the hypersphere is very ideal condition assuming that there exists no intra-class variation. (best viewed in color)}
\label{fig:embed_conv}
\end{figure}

In this paper we explore projections of feature representations onto different hyper-spaces and propose that hypersphere projection has superior performance to linear hyperspace where discriminative analysis and disintegration of multiple classes becomes challenging for networks, Figure~\ref{fig:embed_conv}. We propose that inward scaling applied to projection on hypersphere enhances the network performance in terms of classification and retrieval. Furthermore, we introduce a simpler CNN-based architecture for classification and retrieval tasks and show that non-linear activations (RELU) and techniques like dropout are not necessary for CNN-based networks to generalize. We evaluate proposed network along with inward scaling layer on a number of benchmark datasets for classification and retrieval. We employ MNIST, FashionMNIST\citep{xiao2017fashion}, CIFAR10~\citep{krizhevsky2009learning}, URDU-Characters~\citep{2018:nawaz:DAS} and SVHN~\citep{netzer2011reading} datasets for classification while we employ FashionMNIST for retrieval. Note that the inward scaling layer is not dependent on the proposed network i.e. it can be applied to different types of networks i.e. VGG, Inception-ResNet-V1, GoogleNet~\citep{szegedy2015going} etc and can be trained end-to-end with the underlying network. 
The main contributions of this work are listed as follows. 

\begin{itemize}
  \item [--] We propose the inward scaling layer which can be applied along with the projection layer to ensure maximum separability between divergent classes. We show that the layer enhances the network performance on multiple datasets.
  \item [--] We propose a simpler architecture without dropout~\citep{srivastava2014dropout}, and batch normalization~\citep{ioffe2015batch} layers and experimentally validate that the network achieves results comparable to deeper and wider networks.
  \item [--] We explore the effect of inward scaling layer with different loss functions such as centerloss, contrastive loss and softmax. 
\end{itemize}
The rest of the paper is structured as follows: we explore related literature in Section~\ref{sec:related}, followed by inward scale layer and architecture in Section~\ref{sec:layer}. We review datasets employed and experimental results in Section~\ref{sec:exp}. We finish with conclusion and future work in Section~\ref{sec:conc}.

\section{Related Work}
\label{sec:related}
\subsection{Metric Learning}
Metric learning aims at learning a similarity function, distance metric. Traditionally, metric learning approaches~\citep{weinberger2009distance,ying2012distance,koestinger2012large} focused on learning a similarity matrix \textbf{\sl $M_{i}$}. The similarity matrix is used to measure the similarity between two vectors. Consider feature vectors $X = (x_1, x_2,..., x_n)$ where each vector $x_i$ corresponds to the relevant features. Then the similarity matrix for a corresponding distance metric can be computed as $\|x_i-x_j\| = \sqrt{(x_i-x_j)^TM_i(x_i-x_j)}$ where $x_i$ and $x_j$ are given features. However, in recent metric learning methodologies~\citep{hu2014discriminative,oh2016deep,lu2015multi,hadsell2006dimensionality,2018:nawaz:DAS}, neural networks are employed to learn the discriminative features followed by a distance metric i.e. Euclidean or Manhattan distance $d(x_i, x_j)$ where d is the distance metric used. Contrastive loss~\citep{chopra2005learning,hadsell2006dimensionality} and Triplet loss~\citep{hoffer2015deep,wang2014learning,schroff2015facenet} are commonly used metric learning techniques. Contrastive loss function is a pairwise loss function i.e. reduces the similarity between query and target $L_c(x_i, x_i^{\pm}) = d(x_i, x_i^{\pm})$; where $d$ is the distance metric. However, triplet loss leverages on triplets ($x_i, x_i^{-}, x_i^{+}$) which should be carefully selected to utilize the benefit of the function $L_{t}(x_i, x_i^{-}, x_i^{+}) = d(x_i, x_i^{+}) - d(x_i, x_i^{-}) + \alpha$; where $d(x_i, x_i^{+})$ and $d(x_i, x_i^{-})$ are the distances between query and positive pair and query and negative pair respectively. Note that triplet and pair selection is an expensive process and the space complexity becomes exponential. 

\subsection{Normalization Techniques}
To accelerate the training process of neural networks, normalization was introduced and is still a common operation in modern neural network models. Batch normalization~\citep{ioffe2015batch} was proposed to speed up the training process by reducing the internal covariate shift of immediate features. Scaling and shifting the normalized values becomes necessary to avoid the limitation in representation. The normalization of a layer $L$ can be defined as $\hat{L}^i = \frac{x^i-E[x^i]}{\sqrt{Var[x^i]}}$ where the layer $L$ is normalized along the $i$-th dimension where $x = (x_1, x_2,....,x_n)$ represents the input, $E[x^i]$ represents the mean of activation computed and $Var[x^i]$ represents the variance. The work by~\citep{lecun2012efficient} shows that such normalization aids convergence of the network. Recently, weight normalization~\citep{salimans2016weight} technique was introduced to normalize the weights of convolution layers to speed up the convergence rate.

\subsection{Hypersphere Embedding Techniques}
Different works in literature have explored different hyper-spaces for projection of learned features to figure out manifold with maximum separability between the deep features. Hypersphere embedding is one of the technique where the learned features are projected onto a hypersphere with the $L2$-normalize layer i.e. $\hat{x} = \frac{x}{\|x\|}$. Works in literature have employed hypersphere embedding for different face recognition and verification tasks~\citep{ranjan2017l2,wang2017normface,liu2017sphereface}. These techniques function by imposing discriminative constraints on a hypersphere manifold. As~\citep{ioffe2015batch} explains that scale and shift is necessary to avoid the limitations and are introduced as $y^{(i)} = \gamma^{(i)}\hat{x}^{(i)}+\beta^{(i)}$; where $\gamma$, $\beta$ are learnable parameters. Inspired from this work, techniques such as~\citep{ranjan2017l2} explore $L2$-normalize layer followed by scaling layer which scales the projected features by a factor $\alpha$ i.e. $\frac{\alpha x}{\|x\|}$ where $\alpha$ is the radius of the hypersphere and can be both learnable and predefined, larger values of $\alpha$ result in improved results. However, in~\citep{ranjan2017l2} the $\alpha$ is restricted to the radius of hypersphere and normalizes the features only. Furthermore,~\citep{liu2017sphereface} normalizes the weights of last inner-product layer only and does not explore the scaling factor. The work presented in~\cite{wang2017normface} optimizes both weights and features, and defines the normalization layer as $\|x\|_2 = \sqrt{\sum_ix_i^2+\in}$ without exploring the scaling factor. 

\subsection{Revisiting Softmax-based Techniques}
A generic pipeline for classification tasks consists of a CNN network learning the features of the input coupled with softmax as a supervision signal. We revisit the softmax function by looking at its definition $L_s = -\sum_{i=1}^mlog\frac{e^{W^T_{y_i}x_i+b_{y_{i}}}}{\sum_{j=1}^ne^{W^T_{j}x_i+b_{j}}}$; where $\textbf{x}$ is the learned feature, $W_i \in \mathbb{R}$ denotes weights in the last fully connected layer and $b_i \in \mathbb{R}^n$ is the bias term corresponding to class $\sl i$. By examining, it is clear that $W^T_{i}x_i+b_{i}$ is responsible for the class decision which forms intuition for the necessity of the fully connected layer after normalization.~\citep{liu2017sphereface} reformulates softmax and introduces an angular margin and modifies the decision boundary of softmax as $\|x\|(cosm\theta_{1}-cos\theta_{2}) = 0$ for class 1 and $\|x\|(cos\theta_{1}-cosm\theta_{2}) = 0$ for class 2. This differs from standard softmax in a sense that~\citep{liu2017sphereface} requires $cos(m\theta_{1}) > cos(\theta_{2})$ for the learned feature $\textbf{x}$ to be correctly classified as class 1. This reformulation results in a hypersphere embedding due to the subtended angle. Similarly,~\citep{ranjan2017l2} constraints the softmax by adding a normalization layer. 

\begin{figure}[t!]
\centering
\subcaptionbox{Plot at epoch $3$}{\includegraphics[scale=0.28]{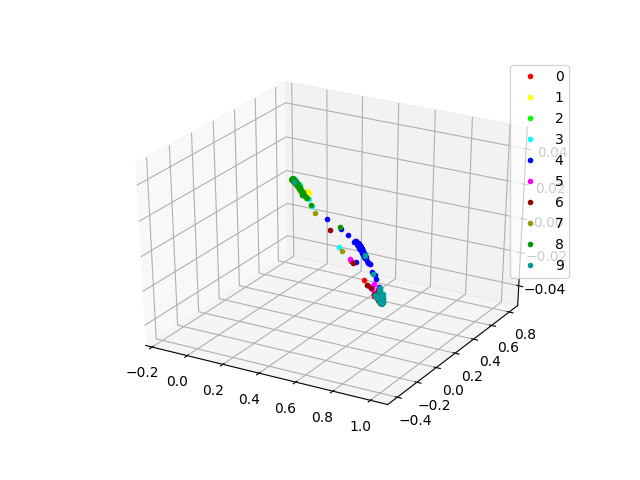}}%
\hfill 
\subcaptionbox{Plot at epoch $15$}{\includegraphics[scale=0.28]{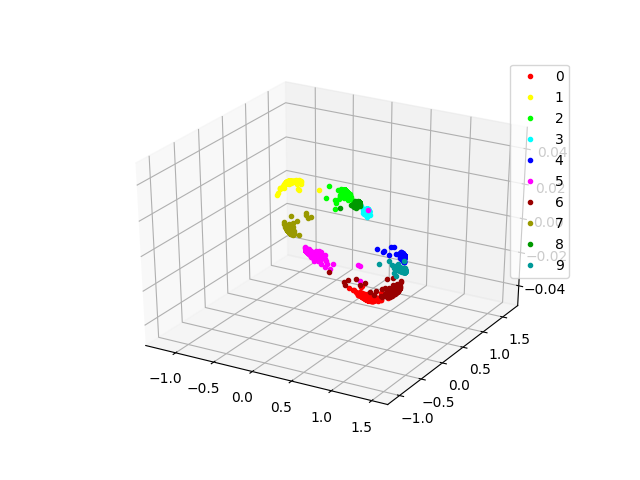}}%
\hfill 
\subcaptionbox{Plot at epoch $30$}{\includegraphics[scale=0.28]{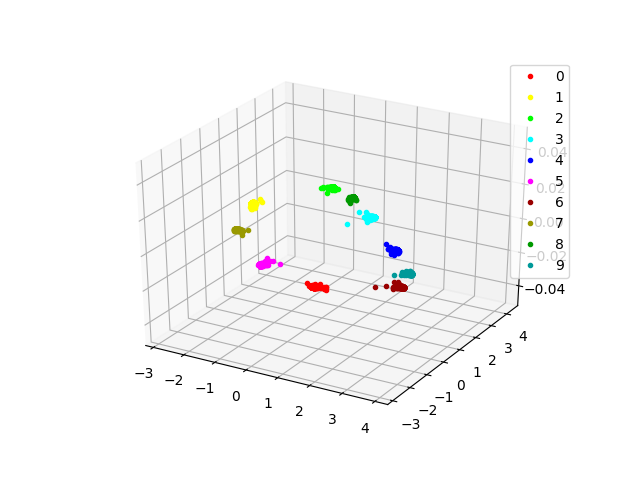}}%
\caption{Plots on test set of MNIST dataset during different epochs. The figure shows realistic plots of test set of MNIST at different epochs. At epoch $3$ the projection of data points on hypersphere embedding space is in initial stages with little to no inward scaling. However, at epoch $15$ effects of inward scaling are visible with the projection being maximum scaled at epoch $30$. (best viewed in color)}
\label{fig:hypersphere-mnist}
\end{figure}

\section{Proposed Method}
\label{sec:layer}
In this section, we explore the intuition behind the inward scale layer and explain why normalization along with a fully connected layer is necessary before the softmax. We term a normalization layer along with the inward scaling factor as the inward scale layer. The reason behind this terminology is that normalization without the inward scaling acts as constraint imposer on the feature space and hampers the discriminative ability of the network. Furthermore, network struggles to converge if either of the layers are removed i.e. normalization, inward scale factor and fully connected. We set some terminology before proceeding with the explanation. 
\begin{table}[h!]
  \begin{center}
    \caption{Some important terminology used throughout this manuscript.}
    \label{tab:termi}
    \begin{tabular}{lSr} 
      \textbf{Terminology} & \textbf{Explanation}\\
      \hline
      $M$ & {Input manifold}\\
      $f(M)$ & {Projected hypersphere manifold}\\
      $x_i$ & {Learned features of class $i$}\\ 
      $W_i$ & {Weight of class $i$}\\
      $b_i$ & {Bias of class $i$}\\
      $\xi$ & {Inward scale factor} \\
      $IS(x, \xi)$ & {Inward scale layer with feature x and scale factor $\xi$} \\
      $FC(W,x,b)$ & {Fully connected layer with weight W, feature x and bias b}\\
    \end{tabular}
  \end{center}
\end{table}

\begin{figure}[t!]
\centering
\subcaptionbox{Plot on test set of MNIST reduced to 2-dimensional features with the softmax as supervision signal without the $IS(x, \xi)$.}{\includegraphics[scale=0.15]{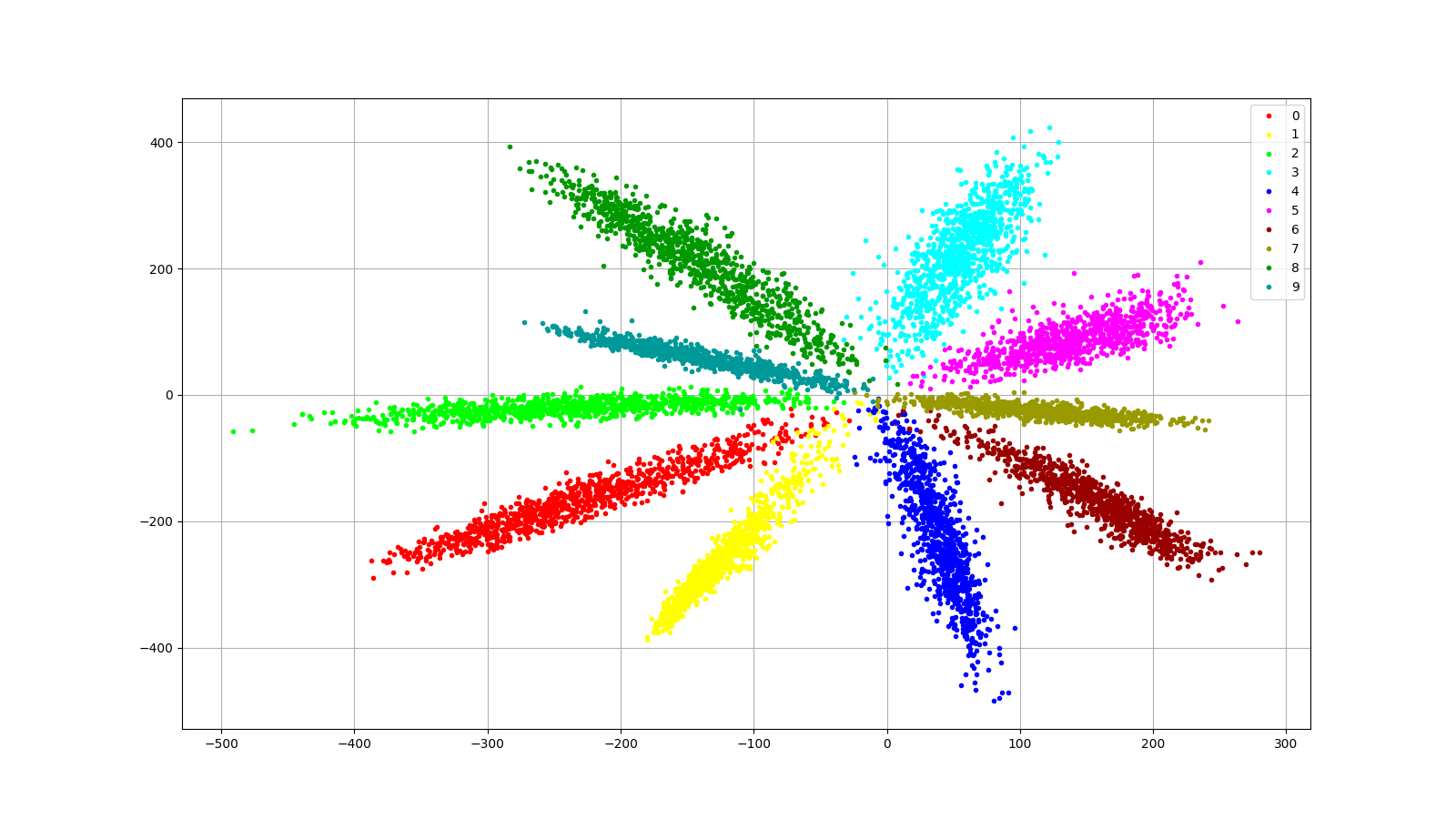}}%
\hfill 
\subcaptionbox{Plot on test set of MNIST reduced to 2-dimensional features with the softmax as supervision signal with the $IS(x, \xi)$.}{\includegraphics[scale=0.11]{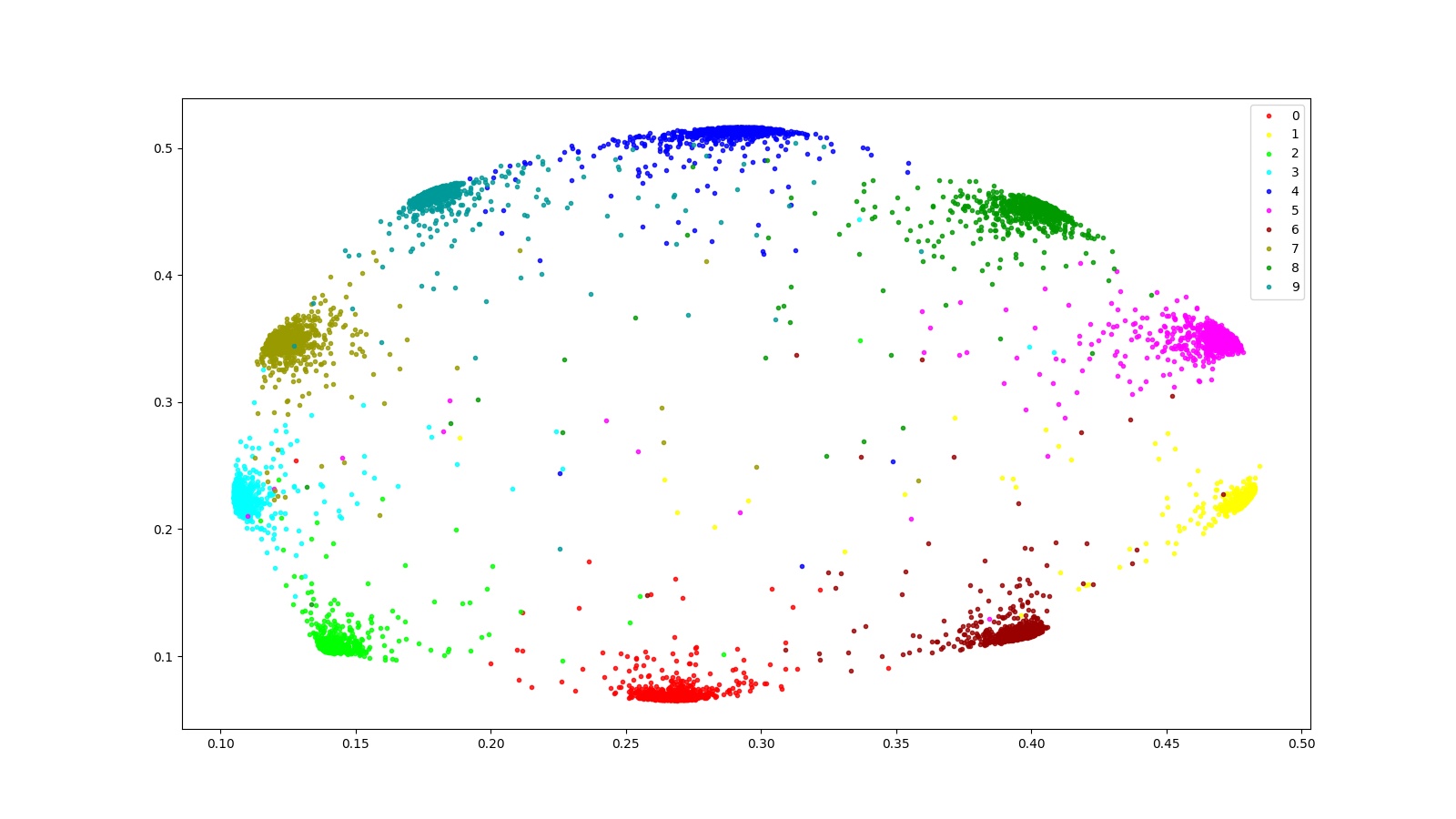}}%
\caption{Comparison of employing softmax with (a) and without (b) the inward scale layer. The softmax tends to have a radial distribution whereas with $IS(x, \xi)$ the distribution changes to hypersphere. Note that the plot (b) has some variation between the features in a radial fashion. This is due to the tendency of softmax. Note that figure (b) is slightly off from the ideal hypersphere embedding, since the features are extracted from the half trained network to establish analogy with the softmax, this scenarios takes place. (best viewed in color)}
\label{fig:comp-softmax}
\end{figure}

The work in~\citep{wang2017normface} establishes that softmax function always encourages well-separated features to have bigger magnitudes resulting in radial distribution Figure~\ref{fig:comp-softmax}(a). However, the effect is minimized in Figure~\ref{fig:comp-softmax}(b) because of the $IS(x, \xi)$. 
\subsection{Inward Scale Layer}
In this paper, we define the inward scale layer as the normalization layer along with the inward scale factor $\xi$. The normalization layer can be defined as in Equation~\ref{eq:norm1}. 

\begin{equation}
\label{eq:norm1}
\hat{x} = \frac{x}{\|x+\mathcal{E}\|}
\end{equation}

where $\mathcal{E}$ is the factor to avoid division by zero. Note that it is unlikely that norm $\|x\| = 0$, but to avoid the risk, we introduce the factor. Inspired from the works in literature~\citep{ranjan2017l2,salimans2016weight} we further introduce a scale factor $\xi$. Unlike employing it in the product fashion as in~\citep{ranjan2017l2}, we couple with the norm in inverse fashion to ensure the scaling of the features as they are projected onto the manifold $f(M)$. In other words, we couple the factor $\xi$ with $\|x\|$ to enhance the norm of the features instead of bounding entire layer. The Equation~\ref{eq:norm2} is modified as $\hat{x} = \frac{x}{\xi(\|x+\mathcal{E}\|)}$. L2-norm can be re-written as $\|x\| = \sqrt{\sum_ix_i^2+\mathcal{E}}$. Thus, $IS(x, \xi)$ can be formulated as follows. 

\begin{equation}
\label{eq:norm2}
\hat{x} = \frac{x}{\xi(\sqrt{\sum_ix_i^2+\mathcal{E}})}
\end{equation}

where $x_i$ is the feature from the previous layer. Note that the factor $\xi$ is not trainable. We experiment with different values of $\xi$ and find that maximum separability is obtained with $\xi = 100$, see appendix A for experiments with different values of $\xi$.

The CNN layers are responsible for providing a meaningful feature space, without the $FC(W,x,b)$ layer, learning non-linear combinations of these features would not be possible. Simply put, the features are classified into different classes due to $FC(W,x,b)$ layers followed by a softmax layer. 
The Figure 3(b) in~\citep{ranjan2017l2} visually illustrates the effect of L2-constrained softmax. On comparing it with our Figure~\ref{fig:hypersphere-mnist}(c) we visually see the effects of the inward scale layer. 
It is necessary to note that we do not modify the softmax and employ it as it is with the $IS(x, \xi)$ which in turn benefits the network with faster convergence and the learned features are discriminative enough for efficient classification and retrieval without the need for any metric learning. 
As the module is fully differentiable and is employed in end-to-end fashion, the gradient with respect to $x_i$ is given as $\frac{\partial L}{\partial x_i}$ and can be solved using the chain-rule, see appendix B for the prove and appendix C for learning curves of the $IS(x, \xi)$.

\subsection{SimpleNet}
Here we explain the proposed network referred to as SimpleNet. The Figure~\ref{fig:arch} represents the architecture visually. Due to the inclusion of the $IS(x, \xi)$ layer, normalizing features or weights during training becomes redundant and adds no performance benefit to the pipeline. So to overcome this redundancy, we do not use any batch or weight normalization layer. Furthermore, it is proposed by~\citep{liu2016large,liu2017sphereface} to remove the ReLU nonlinearity from the networks. We reinforce the idea that ReLU nonlinearity restricts the feature space to non-negative range $[0, +\infty)$ i.e. $\mathbb{R}_{++}$. In order to avoid the feature space from this sufferance, we do not employ ReLU nonlinearity in between the CNN and MaxPool blocks in the network. However, a \textbf{PRelu} layer is added before the last $FC(W,x,b)$ which helps in approximation. It is interesting to note that this does not restrict the feature space to $\mathbb{R}_{++}$. 

\subsection{Activation Maps of SimpleNet and $IS(x, \xi)$}
\begin{figure}[t!]
\centering
\includegraphics[scale=0.44]{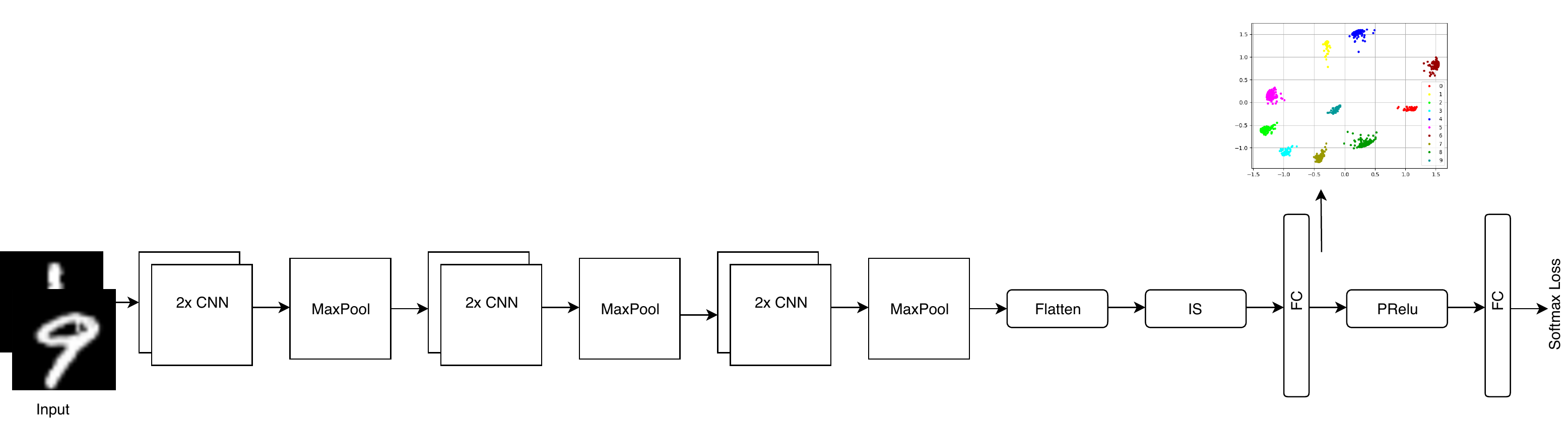}
\caption{The SimpleNet architecture used for the experiments. \textbf{2xCNN} is a block of convolutional neural network (2D) containing two CNN layers followed by a \textbf{MaxPool2D} layer. Whereas the \textbf{IS} is the $IS(x, \xi)$ layer followed by a \textbf{FC} which stands for $FC(W,x,b)$ layer and softmax loss at the end. A single \textbf{PRelu} is used to add non-linearity before the last \textbf{FC} layer. It is worth noting that no batch normalization or data preprocessing is employed throughout the network. (best viewed in color)}
\label{fig:arch}
\end{figure}

\begin{figure}
  \centering
  \begin{center}
  \begin{tabular}{cccccccccc}
  Input Image & Conv1\_1 & Conv2\_2 & Conv3\_3 & $IS(x, \xi)$ Layer\\
  \includegraphics[width=0.15\textwidth]{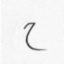} &
  \includegraphics[width=0.15\textwidth]{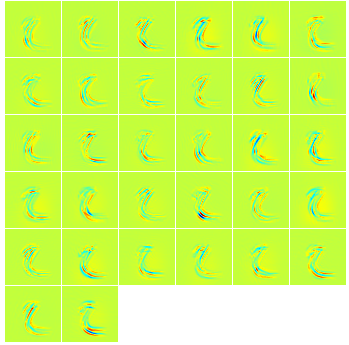} &
  \includegraphics[width=0.15\textwidth]{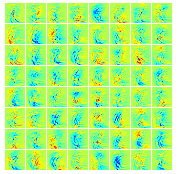} &
    \includegraphics[width=0.15\textwidth]{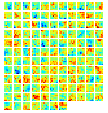} &
     \includegraphics[width=0.15\textwidth]{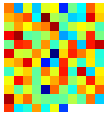} \\
  \end{tabular}
  \end{center}
  \caption{Feature Maps extracted from second convolutional layer from each \textbf{2xCNN} block followed by activation map of the $IS(x, \xi)$ layer. The input image is from the URDU dataset.}
  \label{fig:feature-maps-example}
\end{figure}

\begin{figure}
  \centering
  \begin{center}
  \begin{tabular}{cccccccccc}
  Input Image & Conv\_3 & Pool\_3 & Flatten & $IS(x, \xi)$ Layer\\
  \includegraphics[width=0.15\textwidth]{images/input.jpg} &
  \includegraphics[width=0.15\textwidth]{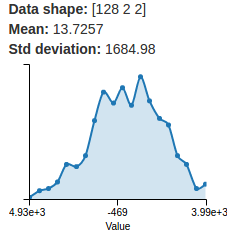} &
  \includegraphics[width=0.15\textwidth]{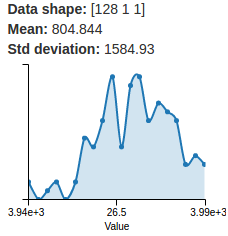} &
    \includegraphics[width=0.15\textwidth]{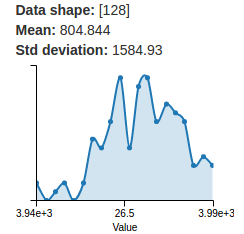} &
     \includegraphics[width=0.15\textwidth]{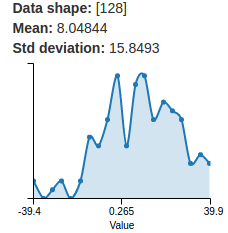} \\
  \end{tabular}
  \end{center}
  \caption{Mean and Standard deviation of data extracted from third convolutional block, maxpool layer, flatten layer and $IS(x, \xi)$ unit from SimpleNet. The input image is from the URDU dataset.}
  \label{fig:meanstddev}
\end{figure}

It becomes intuitive to analyze how the network behaves with the $IS(x, \xi)$ unit in terms of approximating the function. We visualize the activation maps of convolutional layers in the SimpleNet followed by the $IS(x, \xi)$ unit. Figure~\ref{fig:feature-maps-example} is a visual illustration of activation maps extracted from trained SimpleNet~\ref{fig:arch}. Since the scale factor $\xi$ is set to $100$, the change of standard deviation and mean, Figure~\ref{fig:meanstddev}, is according to the factor. Standard deviation is given by $\sqrt{\frac{\sum(x-\hat(x)^2)}{n}}$. With the introduction of the $IS(x, \xi)$ unit, the standard deviation can be re-written as $\frac{1}{\xi}\sqrt{\frac{\sum(x-\hat(x)^2)}{n}}$ and mean of data as $\frac{1}{\xi}(\frac{\sum x}{n})$.


\section{Experimental Results}
\label{sec:exp}
In order to quantify the effects of layer and simplified architecture, in this section we report results of the $IS(x, \xi)$ layer with and without the SimpleNet on multiple datasets. 
\subsection{Experimental Setup}
We perform series of experiments for each dataset. Firstly we report results of different works available in the literature followed by the results of layer with SimpleNet as baseline network and lastly we report results of SimpleNet without the layer. Note that in order demonstrate the modular nature of $IS(x, \xi)$ layer, we perform experiments with different baseline networks containing the proposed layer. The SimpleNet can be trained with standard gradient descent algorithms. In all of the following experiments we employ Adam~\citep{kingma2014adam} optimizer with an initial learning rate of $1e-2$ and employ weight decay strategy to prevent indefinite growing of $\|x\|_2$ because after updating $\|x+\frac{\partial L}{\partial x}\| > \|x\|_2$ for all cases. 
\subsection{Classification Results}
\subsubsection{MNIST and FashionMNIST}
For the basic experiment to quantify results of proposed layer and architecture, we perform the test on MNIST and FashionMNIST dataset which are famous benchmark dataset for neural networks. FashionMNIST is a drop-in replacement for the original MNIST dataset. Table~\ref{tab:mnist} demonstrates the results of $IS(x, \xi)$ layer and SimpleNetwork and compares it with available works in literature.
\begin{table}[h!]
  \begin{center}
    \caption{Accuracy on MNIST and FashionMNIST test set in ($\%$).}
    \label{tab:mnist}
    \begin{tabular}{lSr} 
      \textbf{Methods} & {Dataset}& \textbf{Accuracy ($\%$)}\\
      \hline
      Softmax Loss & MNIST & 98.64\\
       Ours (without $IS(x, \xi)$) & MNIST & 98.40\\
      Ours (with $IS(x, \xi)$) & MNIST & 99.33\\
      Ours (without $IS(x, \xi)$)  & FashionMNIST & 89.64\\
      Ours (with $IS(x, \xi)$) & FashionMNIST & 93.00\\
      \hline
     \cite{ranjan2017l2} & MNIST & 99.05 \\
     ~\cite{zhong2017random}&FashionMNIST&96.35\\
    \end{tabular}
  \end{center}
\end{table}

\subsubsection{CIFAR10 and SVHN}
For the next experiment, we perform the test on CIFAR10 and SVHN datasets. Since MNIST and FashionMNIST are low resolution, grayscale and synthetic datasets, we test the layer on datasets with increasing complexity. Table~\ref{tab:cifar10} demonstrates the results of $IS(x, \xi)$ layer and SimpleNetwork. 
\begin{table}[h!]
  \begin{center}
    \caption{Accuracy on CIFAR10 and SVHN test set in ($\%$). SimpleNet is the baseline network for both experiments.}
    \label{tab:cifar10}
    \begin{tabular}{lSr} 
      \textbf{Methods} & {Dataset} & \textbf{Accuracy ($\%$)}\\
      \hline
      Ours (Without $IS(x, \xi)$) & CIFAR & 58.2\\
      Ours (With $IS(x, \xi)$) & CIFAR & 64.0\\
      Ours (Without $IS(x, \xi)$) & SVHN & 93.20\\
      Ours (With $IS(x, \xi)$) & SVHN & 95.05\\
      \hline
      ~\cite{zagoruyko2016wide}&CIFAR&96.11\\
      ~\cite{zagoruyko2016wide}&SVHN&98.46\\
    \end{tabular}
  \end{center}
\end{table}

\subsubsection{URDU Dataset}
As an experiment on a non-standard dataset, we perform classification on the URDU dataset introduced by~\citep{2018:nawaz:DAS}. Since MNIST and FashionMNIST are $28\times28$, we employ the $64\times64$ format of the URDU dataset to validate the layer on increasing image dimensions along number of channels. Table~\ref{tab:urdu} demonstrates the results of $IS(x, \xi)$ layer. Furthermore, this experiment confirms that the $IS(x, \xi)$ layer is not SimpleNet dependent since URDU dataset is trained using LeNet with softmax as a supervision signal. For comparison, we also experiment with SimpleNet. It is important to note that with networks like GoogleNet, accuracy on the URDU dataset crosses $96\%$ when coupled with the layer. The aim of the experiments is not to demonstrate the superiority of SimpleNet, but to demonstrate the increase in accuracy when a network is coupled with the $IS(x, \xi)$ layer. 
 \begin{table}[h!]
  \begin{center}
    \caption{Accuracy on URDU dataset test set in ($\%$).}
    \label{tab:urdu}
    \begin{tabular}{lrrr} 
      \textbf{Methods} & {Dataset} & Network & \textbf{Accuracy ($\%$)}\\
      \hline
      Ours (Without $IS(x, \xi)$) & URDU & LeNet & 70.02\\
      Ours (With $IS(x, \xi)$) & URDU & LeNet & 71.54\\
      \hline
      Ours (Without $IS(x, \xi)$) & URDU & SimpleNet & 74.03\\
      Ours (With $IS(x, \xi)$) & URDU & SimpleNet & 77.76\\
    \end{tabular}
  \end{center}
\end{table}

\begin{figure}[t!]
\centering
\subcaptionbox{Training loss graph with the $IS(x, \xi)$ layer on CIFAR100 dataset using GoogleNet as baseline network. Classification accuracy is $60.44$.}{\includegraphics[scale=0.15]{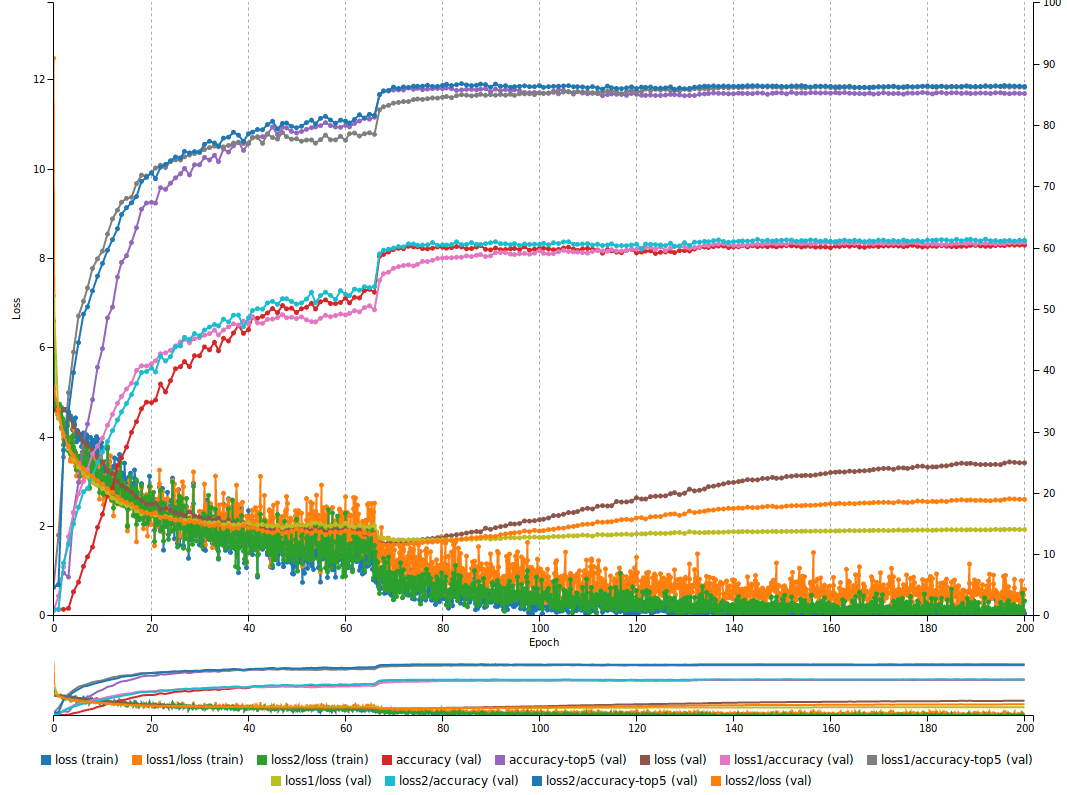}}%
\hfill 
\subcaptionbox{Training loss graph without the $IS(x, \xi)$ layer on CIFAR100 dataset using GoogleNet as baseline network. Classification accuracy is $59.23$.}{\includegraphics[scale=0.15]{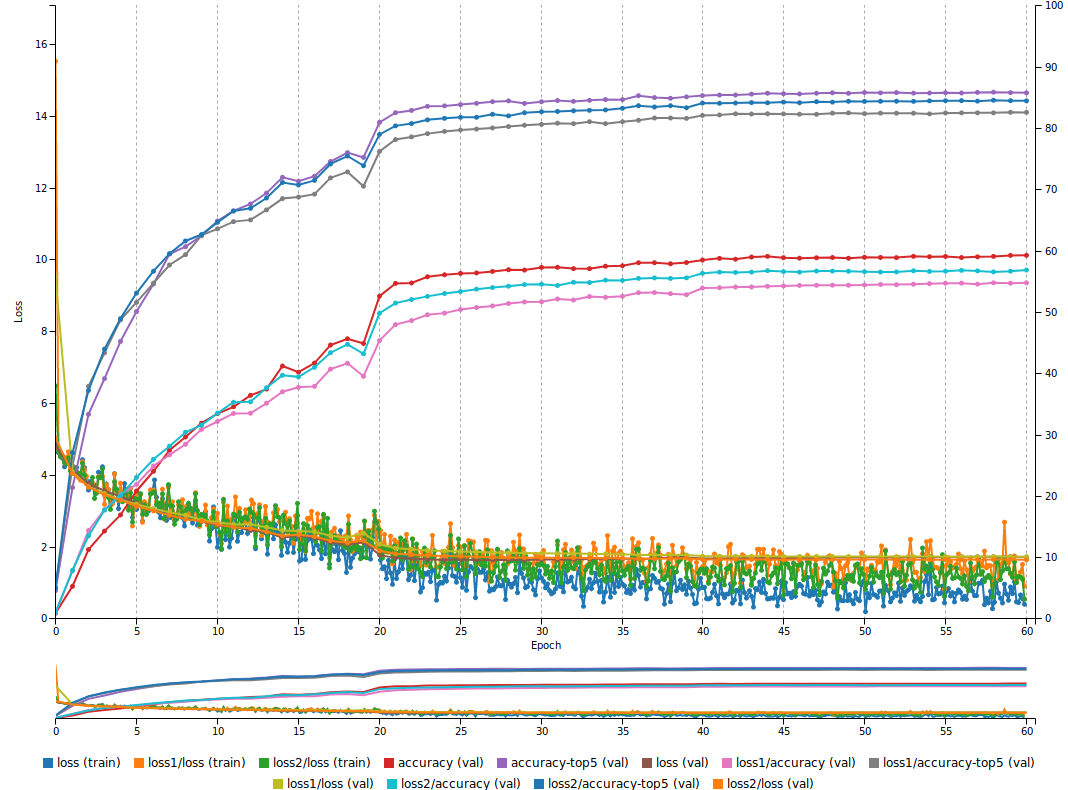}}%
\caption{Plots of training loss on CIFAR100 dataset with and without the proposed layer $IS(x, \xi)$ using GoogleNet as a baseline architecture with no pre or post processing.}
\label{fig:traininggraphcifar100}
\end{figure}

\subsection{CIFAR100}
We perform an additional experiment on the CIFAR100 dataset to confirm the efficacy of the proposed layer. This experiment is particularly interesting because it augments an important claim behind the $IS(x, \xi)$ layer. We employ GoogleNet for this experiment for two reasons: (i) to verify that introduced layer can be coupled with GoogleNet and (ii) CIFAR100 is a large dataset compared to the datasets previously employed, thus, the accuracy with networks like LeNet is not satisfactory. Table~\ref{tab:cifar100} demonstrates the results of GoogleNet on CIFAR100 with and without the $IS(x, \xi)$ layer. Figure~\ref{fig:traininggraphcifar100} visualizes the training graph with and without the proposed unit. It is interesting for readers to note the difference between the two graphs. Note that we imply the idea that projection and scaling happens during each pass and almost simultaneously due to scaling just before the projection. This is the major reason why loss behaves in variating fashion in the start. It should be noted that this does not mean the network struggles to converge. 
\begin{table}[h!]
  \begin{center}
    \caption{Accuracy on CIFAR100 dataset test set in ($\%$). We employ GoogleNet for this experiment.}
    \label{tab:cifar100}
    \begin{tabular}{lrrr} 
      \textbf{Methods} & {Dataset}& \textbf{Accuracy ($\%$)}\\
      \hline
      Ours (Without $IS(x, \xi)$) & CIFAR100  & 59.23\\
      Ours (With $IS(x, \xi)$) & CIFAR100  & 60.44\\
      \hline
      ~\citep{cirecsan2011high} & CIFAR100 & 64.32\\
      ~\citep{goodfellow2013maxout} & CIFAR100 & 65.46\\
      ~\citep{springenberg2014striving} & CIFAR100 & 66.29 \\
      \end{tabular}
  \end{center}
\end{table}

\subsection{Large Scale Classification}
Training state-of-the-art models on ImageNet dataset can take several weeks of computation time. We did not aim for the best performance, rather perform a proof of concept experiment. It is necessary to test if architecture coupled with $IS(x, \xi)$ layer performing best on smaller datasets like CIFAR10, FashionMNIST etc also apply to larger datasets. We employ ILVRC-2012~\citep{ILSVRC15} subset of the ImageNet dataset to train GoogleNet with and without the $IS(x, \xi)$ unit. 

\subsection{Retrieval Results}
In this section we report the retrieval results on FashionMNIST dataset. Most retrieval systems employ Recall@K as a metric to compute the scores. R@K is the percentage of queries in which the ground truth terms are one of the first K retrieved results. To retrieve results, we take query image and simply compute nearest neighbor (euclidean distance) between all images and sort results based on the distance. The first five distances correspond to Recall@K (K = 5) results and so on. We report results for $K = 1,5,10$. Since this is a unimodal retrieval, images are at the input and retrieval end. It is known that Recall@K increases even if one true positive out of Top $K$ is encountered, so the results are almost similar. For a more valid quantitive analysis, we also present results of average occurrence of true positives (TP) in Top $K$. For retrieval, distance minimization is the major objective which softmax alone can not handle efficiently, thus we employ contrastive loss introduced by~\citep{hadsell2006dimensionality} along with softmax for the retrieval problem which shows that the proposed layer $IS(x, \xi)$ can function regardless of the architecture and loss function. 
\begin{table}[h!]
  \begin{center}
    \caption{Recall@K and average occurrence of true positives (TP) in Top $K$ scores for FashionMNIST test set with and without the $IS(x, \xi)$. Note that SimpleNet is the baseline architecture.}
    \label{tab:retres}
    \begin{tabular}{lSr|rr} 
      & \textbf{Without $IS(x, \xi)$} & \textbf{With $IS(x, \xi)$} & \textbf{TP with $IS(x, \xi)$}& \textbf{TP without $IS(x, \xi)$}\\
      \hline
      R@1 &86.75 & 88.75 & 89.74 & 86.70\\
      R@5 & 95.63 & 95.88 & 89.90 & 86.60\\
      R@10 & 97.22 & 97.33 & 90.00 & 85.60\\
    \end{tabular}
  \end{center}
\end{table}

\subsection{Result Discussion}
We explore classification and retrieval tasks with and without the $IS(x, \xi)$ layer. The reported results indicate the superior performance of architecture with the $IS(x, \xi)$ layer. It is important to note that the each experiments is run 5 times and k-fold validation methodology is employed. The architecture with $IS(x, \xi)$ layer maintains the upper bound over its counter part without the $IS(x, \xi)$ layer. In Table~\ref{tab:retres}, the \textbf{TP with} $IS(x, \xi)$ and \textbf{without} $IS(x, \xi)$ indicate the average occurrence of true positives in TopK retrieved results. The reason we employ this metric is Recall@K is incremented even if a single true positive is encountered out of TopK and thus the results with and without $IS(x, \xi)$ layer are almost similar. However, with average true positive, we compute the actual number of true positives out of TopK and compute the average to report more discriminative comparison between the two. Furthermore, we compare the network with state-of-the-art approaches. Note that most of the works obtaining state-of-the-art perform data preprocessing, while we do not employ any pre or post processing technique for any experiment performed and use single optimization policy without fine-tuning hyperparameters for any specific task. 
 
\section{Conclusion and Future Work}
\label{sec:conc}
In this paper, we proposed a novel $IS(x, \xi)$ layer for embedding the learned deep features into an hypersphere. We propose that hypersphere embedding is important for discriminative analysis of the features. We verify the claim with extensive evaluation on multiple classification and retrieval tasks. We propose a simpler architecture for the said tasks and demonstrate that simpler networks achieve results comparable to the deeper networks when coupled with the layer $S(x, \xi)$. Furthermore, the layer module can be added to any network and is fully differentiable and can be trained end-to-end with any network as well. 

In future, we would like to explore different hyperspaces for discriminatively embedding feature representations. Furthermore, we would like to explore constraint-enforced hyperspaces where networks learns a mapping function under certain constraints thus resulting in a desired embedding. 

\bibliography{iclr2019_conference}
\bibliographystyle{iclr2019_conference}
\newpage 

\begin{appendices}
\section{Exploring Different Values of $\xi$}
In this appendix section, we explore different values of the scale factor employed in $IS(x, \xi)$ layer. During training, the hyperspace adapts according to the values of $\xi$. With lower values, stretching is minimum with low intra-class dispersion and results in a hypersphere. Whereas, with greater values of $\xi$, parallel stretch of hyperspace is maximum with compactness of features in one direction. We visually illustrate the effects on MNIST dataset's test set with increasing values of $\xi$ in factor of $10^n$ where $n = 2,3,4,5$. The inward scaling effect is still visible in the figure, but the hypersphere embedding is distorted once the training completes with large false positive rate and low classification accuracy. Note that values of $\xi$ lie in the $10^n; n = 2,3,4,5$ factor, values otherwise yield unsatisfactory results. 

\begin{figure}[t!]
\centering
\subcaptionbox{The value of $\xi$ is $10^2$. Classification accuracy is $99.33$.}{\includegraphics[scale=0.35]{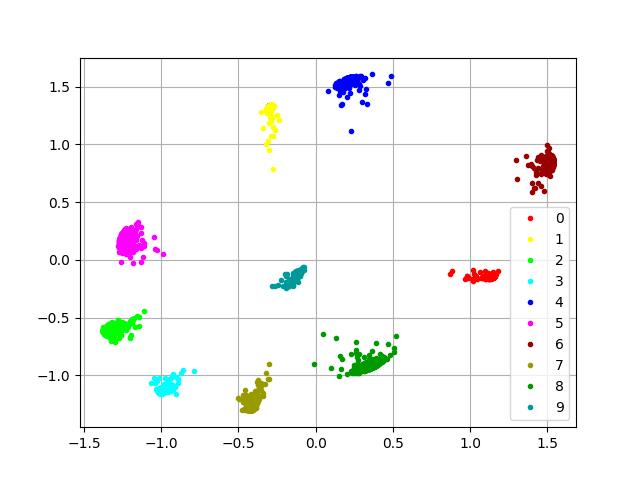}}%
\hfill 
\subcaptionbox{The value of $\xi$ is $10^3$. Classification accuracy is $90.20$.}{\includegraphics[scale=0.35]{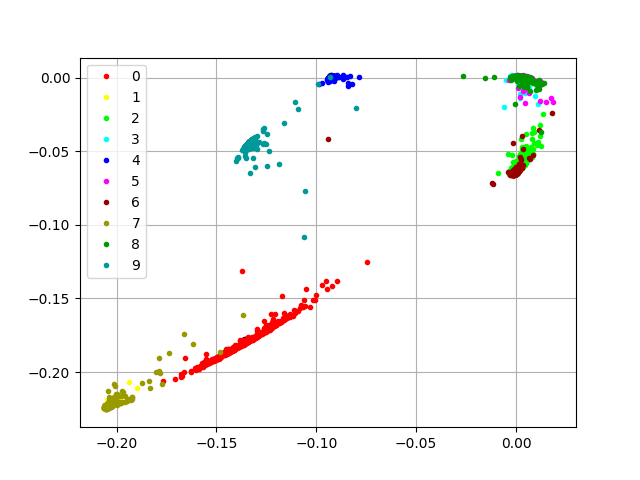}}%
\hfill 
\subcaptionbox{The value of $\xi$ is $10^4$. Classification accuracy is $48.84$.}{\includegraphics[scale=0.35]{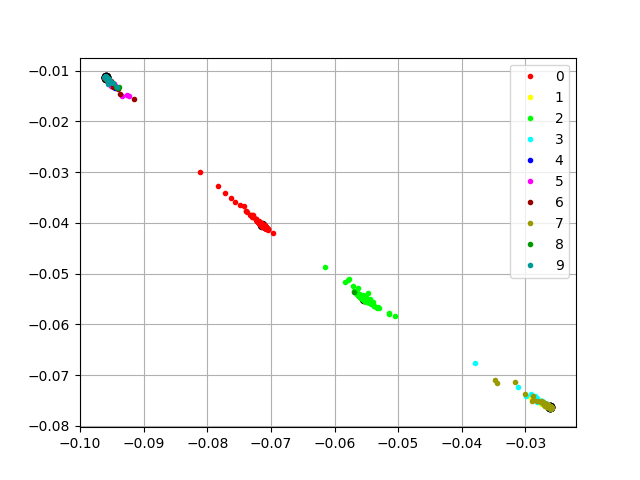}}%
\hfill 
\subcaptionbox{The value of $\xi$ is $10^5$. Classification accuracy is $21.10$.}{\includegraphics[scale=0.35]{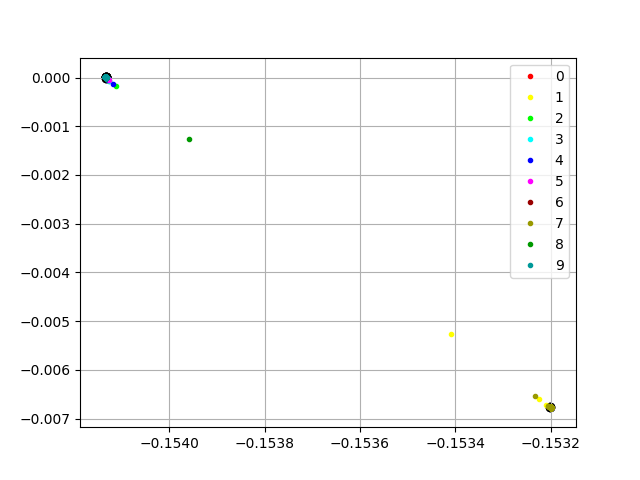}}%
\caption{Plots on MNIST test set with different values of $\xi \in [10^2,10^5]$. The values out of the defined range do not yield satisfactory results due to high compactness of inter-class along one direction. Note that in (c) and (d) concentration at corners starts due to increased inward scaling and stretching to match as the scale changes. It would interesting for the readers to note how the scale changes with values of $\xi$. (Best viewed when zoomed in.) }
\label{fig:scaleval}
\end{figure}

\section{Proving $\frac{\partial L}{\partial x_i}$}
In this appendix section we prove the gradient with respect to $x_i$ i.e. $\frac{\partial L}{\partial x_i}$. We adopt the strategy presented by~\citep{wang2017normface,ranjan2017l2}. Since $\xi$ is not a learnable parameter, we can ignore it during gradient computation. We know from Equation~\ref{eq:norm1} that $\hat{x} = \frac{x}{\|x+\mathcal{E}\|}$, similarly $\hat{x} = \frac{x}{\xi(\sqrt{\sum_ix_i^2+\mathcal{E}})}$ where $\|x\|_2 = \sqrt{\sum_ix_i^2+\mathcal{E}}$. We have $\frac{\partial L}{\partial x_i}$ as follows. $$\frac{\partial L}{\partial x_i}  = \frac{\partial L}{\partial \hat{x}_i}\frac{\partial \hat{x}_i}{\partial x_i} + \gamma$$ where $\gamma = \sum_j\frac{\partial L}{\partial \hat{x}_j}\frac{\partial \hat{x}_j}{\partial \|x\|_2}\frac{\partial \|x\|_2}{\partial x_i}$. The gradient of $\gamma$ is denoted as $\hat{\gamma} = -\frac{x_i}{\|x\|_2}\sum\frac{\partial L}{\partial \hat{x}_j}\frac{\hat{x}_j}{\|x\|_2^2}$. We ignore $\hat{\gamma}$ in the final equation because our main objective is to demonstrate gradient w.r.t to the introduced $IS(x, \xi)$. Using these notations, we proceed as follows. $$\frac{\partial L}{\partial \hat{x}_i}\frac{\partial \hat{x}_i}{\partial x_i} = \frac{\partial L}{\partial \hat{x}}\frac{1}{\|x\|_2}$$. $$\frac{\partial L}{\partial x_i} = \frac{\frac{\partial L}{\partial \hat{x_i}} - \hat{\gamma}}{\|x\|_2}$$

\section{Learning Curves of $IS(x, \xi)$}
For more intuitive understanding of the proposed layer $IS(x, \xi)$ it is important to visualize the training graph plotted as the loss decreases. It would be interesting to note the difference between the graph with and without the $IS(x, \xi)$ layer. For the sake of employing different networks with the layer, this plot is with LeNet. 
\begin{figure}[t!]
\centering
\subcaptionbox{Training loss graph with the $IS(x, \xi)$ layer on URDU dataset using LeNet as baseline network. Classification accuracy is $71.54$.}{\includegraphics[scale=0.15]{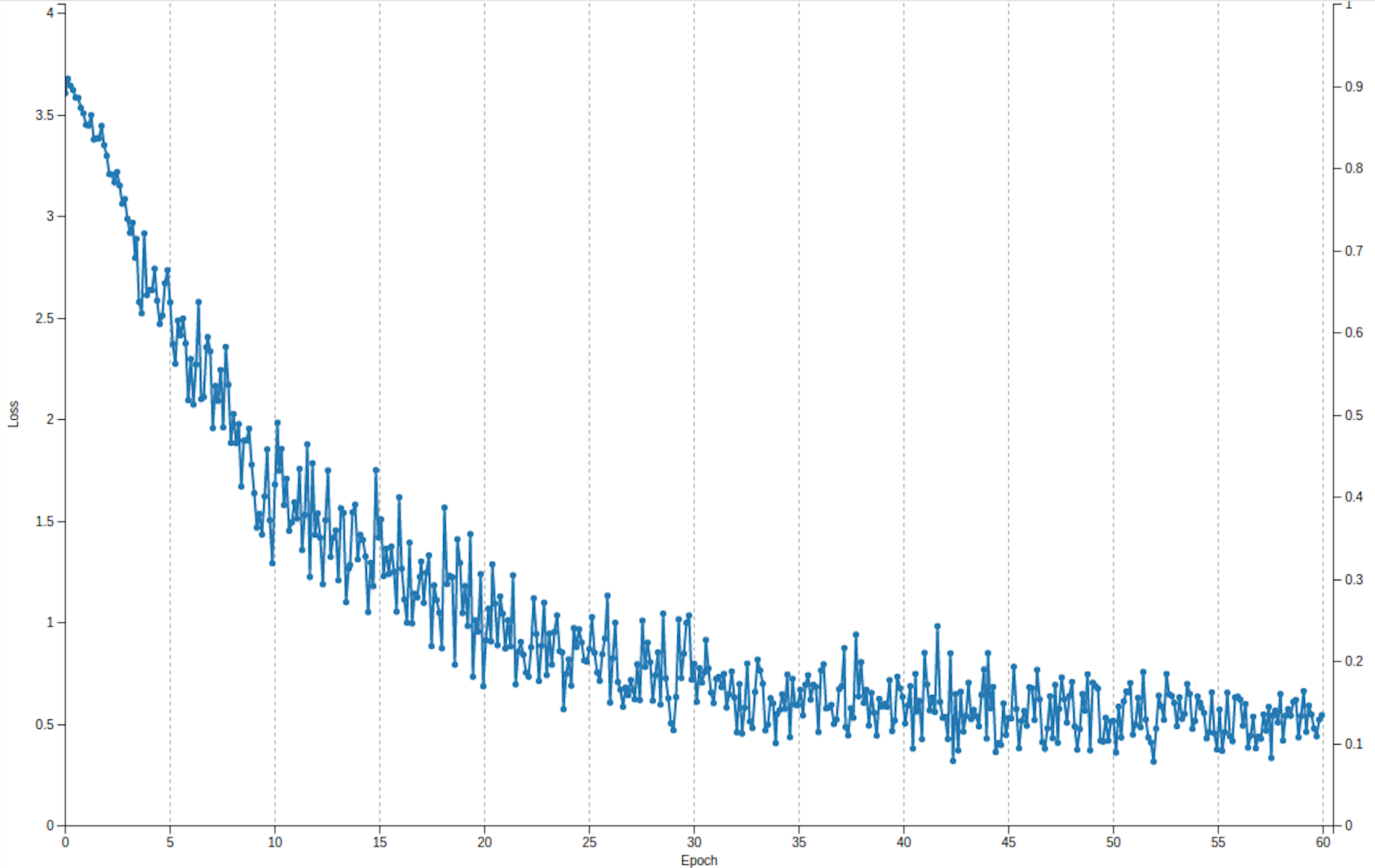}}%
\hfill 
\subcaptionbox{Training loss graph without the $IS(x, \xi)$ layer on URDU dataset using LeNet as baseline network. Classification accuracy is $70.02$.}{\includegraphics[scale=0.15]{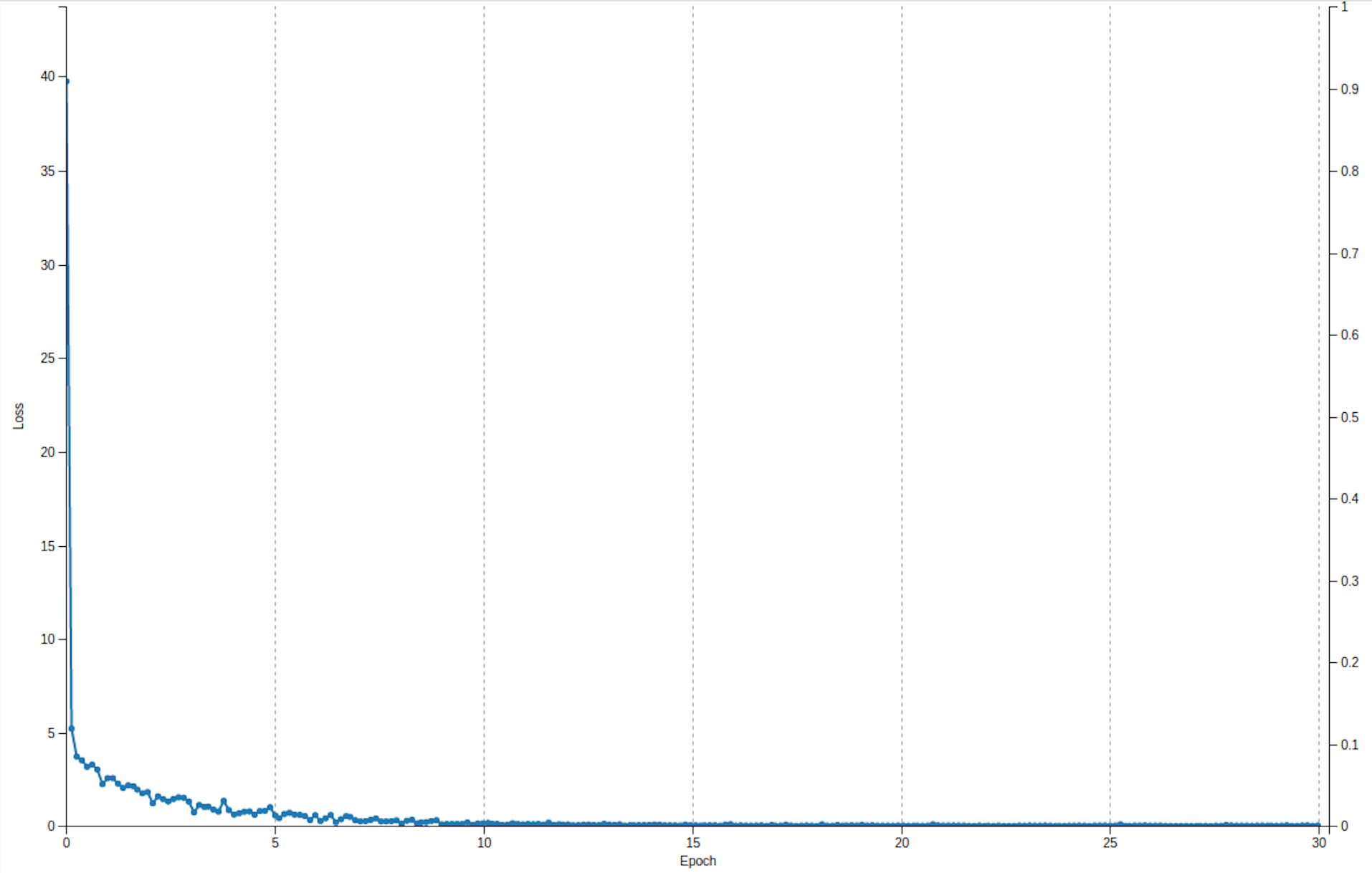}}%
\caption{Plots of training loss on URDU dataset with and without the proposed layer $IS(x, \xi)$ using LeNet as a baseline architecture. }
\label{fig:traininggraph}
\end{figure}

\end{appendices}

\end{document}